%% file: final.tex
\documentclass[10pt,twocolumn,letterpaper]{article}

\input{Packages.tex}

\usepackage[breaklinks=true,bookmarks=false]{hyperref}

\iccvfinalcopy 

\input{doc/0PaperId.tex}

\ificcvfinal\fi

\begin{document}

\input{doc/1Title.tex}

\maketitle
\ificcvfinal\thispagestyle{empty}\fi

\input{doc/2Abstract.tex}

\input{doc/3Introduction.tex}
\input{doc/4RelatedWork.tex}

\input{doc/5Approach.tex}
\input{doc/6Experiments.tex}
\input{doc/7Conclusion.tex}
\input{doc/8Acknowledgement.tex}

{\small
\bibliographystyle{ieee}
\bibliography{final.bbl}
}

\end{document}

%% file: Packages.tex
\usepackage{iccv}
\usepackage{times}
\usepackage{epsfig}
\usepackage{graphicx}
\usepackage{amsmath}
\usepackage{amssymb}

\usepackage{subfigure}
\usepackage{tabularx}
\usepackage{scrextend}

%% file: doc/0PaperId.tex
\def\iccvPaperID{0009} 

%% file: doc/1Title.tex
\title{Fast Visual Object Tracking with Rotated Bounding Boxes}

\author{Bao Xin Chen\qquad John K. Tsotsos\\
Department of Electrical Engineering and Computer Science, and Centre for Vision Research\\
York University\\
Toronto, Canada\\
{\tt\small \{baoxchen, tsotsos\}@eecs.yorku.ca}
}

%% file: doc/2Abstract.tex
\begin{abstract}
In this paper, we demonstrate a novel algorithm that uses \textbf{ellipse fitting} to estimate the bounding box rotation angle and size with the segmentation(mask) on the target for online and real-time visual object tracking. Our method, \textbf{SiamMask\_E}, improves the bounding box fitting procedure of the state-of-the-art object tracking algorithm SiamMask and still retains a fast-tracking frame rate (\textbf{80 fps}) on a system equipped with GPU (GeForce GTX 1080 Ti or higher). We tested our approach on the visual object tracking datasets (VOT2016, VOT2018, and VOT2019) that were labeled with rotated bounding boxes. By comparing with the original SiamMask, we achieved an improved Accuracy of \textbf{65.2\%} and \textbf{30.9\%} EAO on VOT2019, which is 5.6\% and 2.6\% higher than the original SiamMask.\ificcvfinal The implementation is available on GitHub: \url{https://github.com/baoxinchen/siammask_e}.\fi
\end{abstract}

%% file: doc/3Introduction.tex
\section{Introduction}
\label{Introduction}

\begin{figure}[t]
\newcommand{\inputimagewidth}{1.9cm}
\setlength{\tabcolsep}{0.15em} 
\centering
\begin{tabular}{p{2.2cm}ccc}
\includegraphics[width = \inputimagewidth]{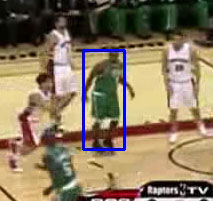}&
\includegraphics[width = \inputimagewidth]{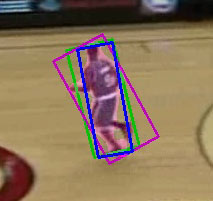}&
\includegraphics[width = \inputimagewidth]{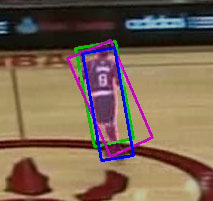}&
\includegraphics[width = \inputimagewidth]{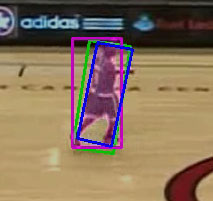}\\
\includegraphics[width = \inputimagewidth]{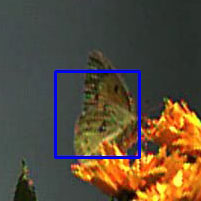}&
\includegraphics[width = \inputimagewidth]{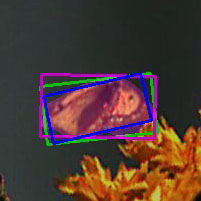}&
\includegraphics[width = \inputimagewidth]{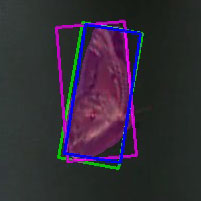}&
\includegraphics[width = \inputimagewidth]{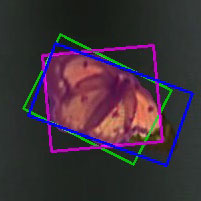}\\
\includegraphics[width = \inputimagewidth]{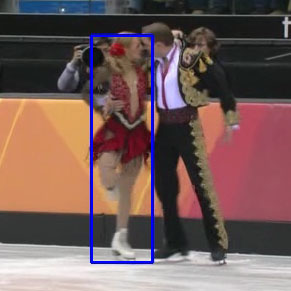}&
\includegraphics[width = \inputimagewidth]{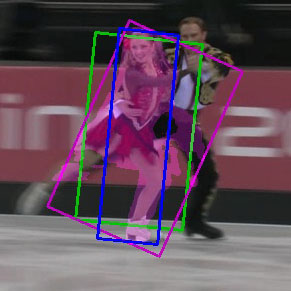}&
\includegraphics[width = \inputimagewidth]{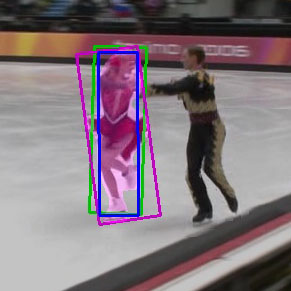}&
\includegraphics[width = \inputimagewidth]{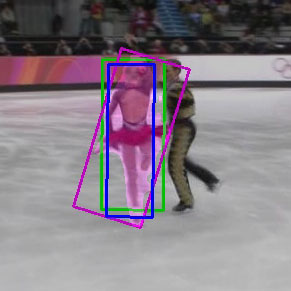}\\
\includegraphics[width = \inputimagewidth]{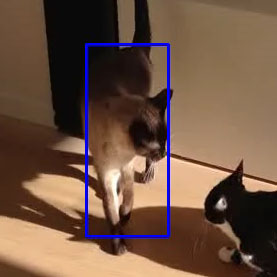}&
\includegraphics[width = \inputimagewidth]{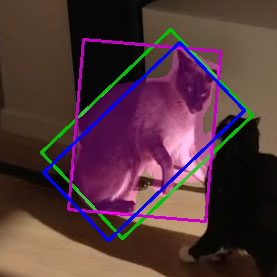}&
\includegraphics[width = \inputimagewidth]{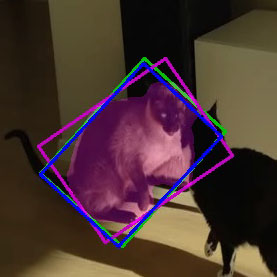}&
\includegraphics[width = \inputimagewidth]{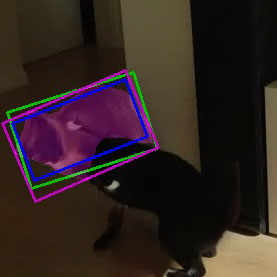}\\
\centering 
Init & \multicolumn{3}{c}{Output}\\
\end{tabular}
\vspace{0.2cm}
\caption{Our approach SiamMask\_E yields lager IoU between the ground truth (blue) and its prediction (green) than the original SiamMask (magenta). SiamMask\_E predicts a higher accuracy on the orientation of the bounding boxes which improves the average overlap accuracy (A) and expected average overlap (EAO).}
\label{fig:siammask_vs_siammaske}
\end{figure}

Visual object tracking is an important element of many applications such as person-following robots (\ificcvfinal\cite{chen2017personfollowing}~\cite{chen2017integrating}~\fi\cite{ren2016real}~\cite{koide2018convolutional}), self-driving cars (\cite{agarwal2018study}~\cite{cho2014multi}~\cite{petrovskaya2009model}~\cite{buyval2018realtime}), or surveillance cameras (\cite{gajjar2017human}~\cite{lee2017online}~\cite{zhou2016moving}~\cite{xu2018real}), etc. The performance of such systems critically depends on a reliable and efficient object tracking algorithm. It is especially important to track an object \textbf{online} and in \textbf{real-time} when the camera is running under challenging situations: illumination, changing pose, motion blurring, partial and full occlusion, etc. These two fundamental features are the core requirements for human-robot interactions (e.g., person-following robots).

To address the visual object tracking problems, many benchmarks have been developed, such as Object Tracking Benchmark (OTB50~\cite{WuLimYang13} and OTB100~\cite{WuLimYang15}), and Visual Object Tracking Challenges (VOT2016~\cite{Kristan2016a}, VOT2018~\cite{Kristan2018a}, VOT2019~\cite{Kristan2019a}). 
In OTB datasets, ground truth was labeled by axis aligned bounding boxes and while in VOT datasets rotated bounding boxes were used. 
Comparing between axis-aligned bounding boxes and rotated bounding boxes, rotated bounding boxes contain a minimal amount of background pixels~\cite{Kristan2016a}. Thus, the datasets with rotated bounding boxes have the tighter enclosed boxes than the axis-aligned bounding boxes. As well as, the rotated bounding boxes provide the object orientation in the image plane. The orientation information can be further used to solve many computer vision problems (e.g., action classification).

Despite the advantage of rotated bounding boxes, it is very computationally intensive to estimate the rotation angle and scale of the bounding boxes. Many researchers have developed novel algorithms to settle the problem. But most of them have limitations in terms of tracking speed or accuracy \cite{hua2015online}, \cite{rout2018rotation}. In the meantime, fully convolutional Siamese networks~\cite{bertinetto2016fully} had become popular in the field of object tracking. However, the original Siamese networks did not solve the rotation problem. Wang \textit{et al.} (SiamMask) \cite{wang2018fast} have been inspired by the advanced version of Siamese network (SiamRPN~\cite{li2018high}, SiamRPN++~\cite{li2018siamrpn++}) and wide range of image datasets (Youtube-VOS~\cite{xu2018youtube}, COCO~\cite{lin2014microsoft}, ImageNet~\cite{russakovsky2015imagenet}, etc.). SiamMask is able to predict a segmentation mask on the target for tracking and fits a minimum area rotated bounding box in real-time (87 fps). 

In this paper, we propose a novel efficient rotated bounding box estimation algorithm when a segmentation/mask of an object is given. Particularly, the masks are generated by SiamMask. The key problem is to predict the rotation angle of the bounding boxes. Inspired by the conic fitting problem described by Fitzgibbon \textit{et al.}~\cite{fitzgibbon1996buyer}, we try to fit an ellipse on the mask to compute the rotation angle. Once the rotation angle is known, then we could fit a rotated rectangle on the mask. Our algorithm consists of two parts: (1) rotation angle estimation, and (2) scale calculation. Details will be provided in Section~\ref{approach}.

The contribution of this paper can be summarized in the following three aspects: 
\begin{enumerate}
  \item a new real-time state-of-the-art \textbf{object tracking algorithm} on the datasets that are labeled with rotated bounding boxes, e.g., VOT challenge series (2015-2019)~\footnote{\url{http://www.votchallenge.net/challenges.html}}.
  \item a fast novel \textbf{rotated bounding box estimation algorithm} when a segmentation/mask is given. This algorithm can be used to generate rotated bounding box ground truth from any segmentation datasets to train a rotation angle regression model. This is the main contribution of the paper.
  \item the source code~\ificcvfinal\footnote{\url{https://github.com/baoxinchen/siammask_e}} \fi will be released as an additional package to PySOT~\footnote{\url{https://github.com/STVIR/pysot}} which is written by SenseTime Video Intelligence Research team.
\end{enumerate}

The paper is structured as follows. The most relevant work will be briefly summarized in Section~\ref{relatedwork}. Then, we will describe our approach in detail in Section~\ref{approach}. The evaluation of the algorithm is in Section~\ref{experiments}. Finally, Section~\ref{conclusion} concludes the paper and discusses future work. 

%% file: doc/4RelatedWork.tex
\section{Related Work}
\label{relatedwork}

In this section, we discuss the history of the Siamese network based tracking algorithms and several trackers that yield rotated bounding boxes.

\subsection{Siamese network based trackers}

The first Siamese network based object tracking algorithm (\textbf{SiamFC}) was introduced by Bertinetto \textit{et al.}~\cite{bertinetto2016fully} in 2016. The Siamese network is trained offline on a dataset for object detection in videos. The input to the network are two images, one is an exemplar image $z$, the other one is the search image $x$. Then, a dense response map is generated from the output of the network. \textbf{SiamFC} learns and predicts the similarity between the regions in $x$ and the exemplar image $z$. In order to handle the object scale variantion, SiamFC searches for objects at five scales $1.025^{\{−2,−1,0,1,2\}}$ near the target's previous location. As a result, there will be 5 forward passes on each frame. SiamFC runs at about 58 fps, which is the fastest fully convolutional network (CNN) based tracker comparing to online training and updating networks in 2016. However, SiamFC is an axis-aligned bounding box tracker. It couldn't outperform the online training and updating deep CNN tracker MDNet~\cite{nam2016learning} (1 fps) in terms of average overlap accuracy.

He \textit{et al.}~\cite{he2018twofold} combines two branches (\underline{S}emantic net and \underline{A}ppearance net) of Siamese network (\textbf{SA-Siam}) to improve the generalization capability of \textbf{SiamFC}. Two branches are individually trained, and then the two branches are combined to output the similarity score at testing. S-Net is an AlexNet~\cite{krizhevsky2012imagenet} pretrained on an image classification dataset. A-Net is a SiamFC pretrained on an object detection from video dataset. S-Net improves the discrimination power of the SA-Siam tracker because different objects activate different sets of feature channels in the Semantic branch. Due to the complexity of the two branches, SA-Siam runs at 50 fps when tracking with pretrained model.

By modifying the original Siamese net with a Region Proposal Network(RPN)~\cite{ren2015faster}, Li \textit{et al.}~\cite{li2018high} proposed a Siamese Region Proposal Network (\textbf{SiamRPN}) to estimate the target location with the variable bounding boxes. The output of \textbf{SiamRPN} contains a set of anchor boxes with corresponding scores. So, the bounding box with the best score is considered as the target location. The benefit of RPN is to reduce the multi-scale testing complexity in the traditional Siamese networks (SiamFC, SA-Siam). An updated version SiamRPN++~\cite{li2018siamrpn++} has released in 2019. In terms of processing speed, SiamRPN is 160 fps and SiamRPN++ is about 35 fps.

Unlike \textbf{SiamFC}, \textbf{SA-Siam}, and \textbf{SiamRPN} yielding axis-aligned bounding boxes, \textbf{SiamMask}~\cite{wang2018fast} uses the advantage from a video object segmentation dataset and trained a Siamese net to predict a set of masks and bounding boxes on the target. The bounding boxes are estimated based on the masks using rotated minimum bounding rectangle (MBR) at the speed of 87 fps. However, the MBR does not always predict the bounding boxes that perfectly align with the ground truth bounding boxes (see Figure~\ref{fig:siammask_vs_siammaske}). Although the same bounding boxes prediction algorithm used in VOT2016 for generating the ground truth can improve the average overlap accuracy dramatically, the running speed decreases to 5 fps. To address this problem, we present a new method in Section~\ref{approach} that can process frames in real-time and achieves a better result.

\subsection{Rotated bounding boxes}

Beside the Siamese network trackers, Nebehay \textit{et al.}~\cite{nebehay2014consensus} (CMT) use a key-point matching approach to scale and rotate the bounding boxes. But, this tracker cannot handle deformable objects. \cite{nebehay2015clustering} is an update of CMT, and the processing speed dropped to 11 fps. 

Hua \textit{et al.}~\cite{hua2015online} suggest a proposal selection method (optical flow~\cite{brox2010large} and Hough transform~\cite{hough1962method}) to filter out a group of locations and orientations that very likely contains the object. Then, they use three cues (detection confidence, objectness measures from object edges and motion boundaries) to determine which location has the highest likelihood. But, this approach also couldn't run in real-time (0.3 fps). 

Zhang \textit{et al.}~\cite{zhang2015joint} propose a rotation estimation method using Log-Polar transformation. In Log-Polar coordinate, a set of 36 rotation sample are chosen on every $\Delta=\frac{2\pi}{R}$, where $R=36$. But, the rotation sample set also increases the rum-time of KCF~\cite{henriques2014high} tracker by 36 times.

Guo \textit{et al.}~\cite{guo2017structure} build a structure-regularized compressive tracking (SCT) with online update. During the detection stage, SCT samples several candidates with different rotation angles based on integral image and quadtree segmentation. SCT runs on a computer system without GPU at 15 fps. 

Recently, a rotation adaptive tracking approach was introduced by Rout \textit{et al.}~\cite{rout2018rotation}. The authors assume that the rotation angle is limited within a range (e.g., $\pm10^\circ$). However, this assumption doesn't always hold. He \textit{et al.}~\cite{he2018towards} built on top of SA-Siam~\cite{he2018twofold} with angle estimation strategy. Although the method could reduce the processing time, it still limits the rotation angle to some degrees (e.g., $-\pi/8$, $\pi/8$). In order to find an arbitrary rotation angle, we present our approach in the next section.

%% file: doc/5Approach.tex
\section{Approach}
\label{approach}

\begin{figure}[t]
\newcommand{\inputimagewidth}{2.2cm}
\begin{center}
\subfigure[]{\includegraphics[width = \inputimagewidth]{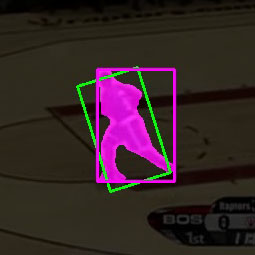}}
\subfigure[]{\includegraphics[width = \inputimagewidth]{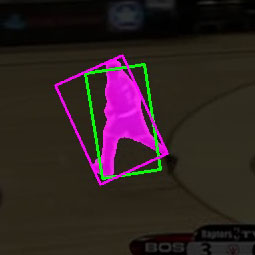}}
\subfigure[]{\includegraphics[width = \inputimagewidth]{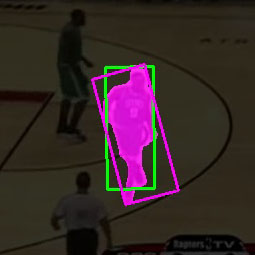}}
\end{center}
\caption{This figure shows some examples of minimum area rectangle (magenta); this does not determine bounding boxes according to the geometric shape and point distribution of the segmentation/mask. Thus, the rotation angles are not as accurate as our approach (green)}
\label{fig:geometric_shape}
\end{figure}

\begin{figure*}[t]
\newcommand{\inputimagewidth}{2.2cm}
\begin{center}
\subfigure[input]{\includegraphics[width = \inputimagewidth]{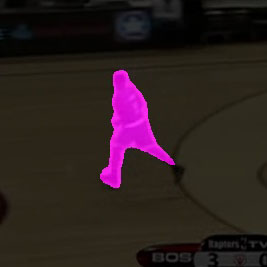}}
\subfigure[fit an ellipse]{\includegraphics[width = \inputimagewidth]{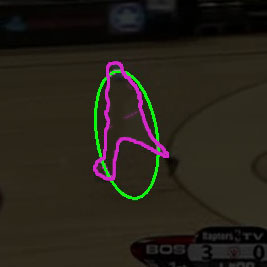}}
\subfigure[transform]{\includegraphics[width = \inputimagewidth]{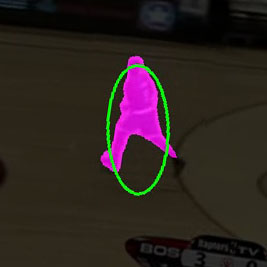}}
\subfigure[ellipse to box]{\includegraphics[width = \inputimagewidth]{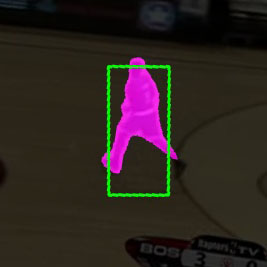}}
\subfigure[min-max(blue)]{\includegraphics[width = \inputimagewidth]{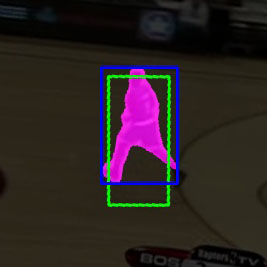}}
\subfigure[intersection]{\includegraphics[width = \inputimagewidth]{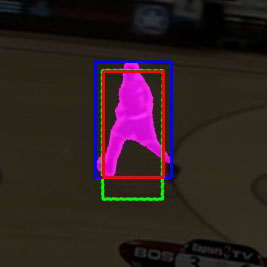}}
\subfigure[inverse transformation]{\includegraphics[width = \inputimagewidth]{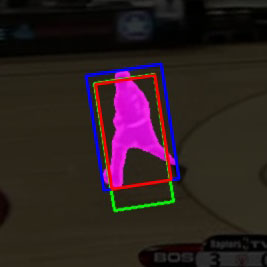}}
\end{center}
\caption{Our algorithm includes seven steps: (a) take a target mask as input. (b) apply an ellipse fitting algorithm~\cite{fitzgibbon1996buyer} on edge of the mask (Here, we have the points on the edge as a set $A$ in Equation~\ref{eq:minimizing}), then determine the center of the ellipse and the rotation angle. (c) compute the affine transformation matrix using the rotation angle and the center from the ellipse, then apply the transformation on the ellipse center. (d) apply a rectangular rotated bounding box (green) on the ellipse. (e) draw a min-max axis-aligned bounding box (blue) on the transformed mask. (f) calculate the intersection of the blue box and green box to form a new bounding box (red). (g) calculate the inverse of the affine transformation matrix, then apply transformation to convert back to the original image coordinate, and output the red box.}
\label{fig:steps}
\end{figure*}

In the original SiamMask~\cite{wang2018fast} tracker, Wang et al. compared three different bounding boxes estimation algorithms: min-max axis-aligned rectangle (\textit{Min-max}), minimum area rectangle (\textit{MBR}), and optimal bounding box~\cite{Kristan2016a} (\textit{Opt}). Due to the computational burden, \textit{Opt} could not perform in real-time (5fps). SiamMask with \textit{MBR} is the real-time (87 fps) state-of-the-art tracker in terms of average overlap Accuracy. Although \textit{MBR} performs better than the other bounding box estimation algorithms, it has a weakness such that minimum area rectangle could not represent the geometric shape and point distribution of the masks (see Figure~\ref{fig:geometric_shape}). As a result, most of the estimated bounding boxes are not in the correct orientation. In the following subsections, we will discuss an alternate solution to generate bounding boxes with correct rotation angle and tighter size by post-processing on the output mask from SiamMask. Our method consists of the steps in Figure~\ref{fig:steps}.

\subsection{Rotation angle estimation}

To estimate the rotation angle, we adopted the fitEllipse API provided by OpenCV3~\footnote{\url{https://opencv.org/}} which use a least-squares scheme~\cite{gander1980least} to solved the ellipse fitting problem. An improved version was described in \cite{halir1998numerically}. This algorithm (B2AC) Algebraic distance with quadratic constraint was first introduced by Fitzgibbon \textit{et al.}~\cite{fitzgibbon1996buyer}. 

\begin{figure}[t]
\newcommand{\inputimagewidth}{5cm}
\begin{center}
\includegraphics[width = \inputimagewidth]{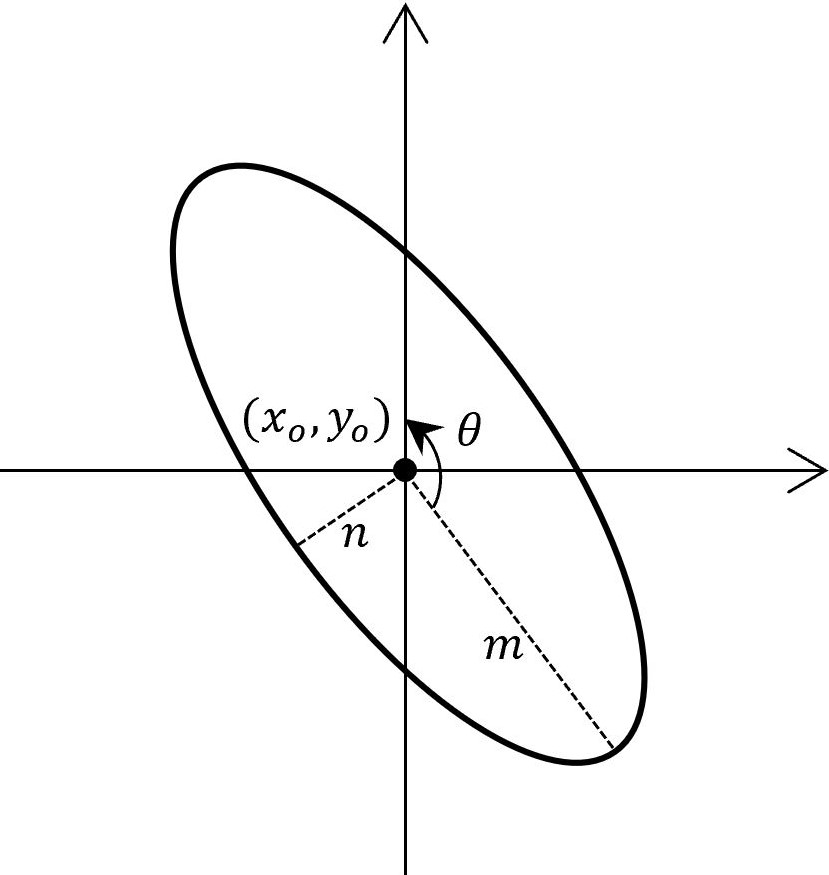}
\end{center}
\caption{Ellipse notations}
\label{fig:ellipse_notaions}
\end{figure}

An ellipse can be formulated using a conic equation with a constraint:
\begin{equation}
\label{eq:conic}
\begin{aligned}
F(x,y) = &ax^2 + bxy + cy^2 + dx + ey + f = 0\\
         &\text{where, } b^2 - 4ac < 0
\end{aligned}
\end{equation}
In Equation~\ref{eq:conic}, $a,b,c,d,e,f$ are the coefficients of the ellipse and $x,y$ are the points on the ellipse. By grouping the coefficients into a vector, we have the following two vectors:
\begin{equation}
\label{eq:parameters}
\begin{aligned}
&\textbf{a} = [a, b, c, d, e, f]^T\\
&\textbf{x} = [x^2, xy, y^2, x, y, 1]
\end{aligned}
\end{equation}
So, the conic can be written as:
\begin{equation}
\label{eq:conic_vector}
\begin{aligned}
F(\textbf{x}) = \textbf{x} \cdot \textbf{a} = 0
\end{aligned}
\end{equation}
To fit an ellipse on a set of points $A=\{(x_1, y_1),...,(x_N, y_N)\}, where |A|=N$, we need to find the coefficient vector \textbf{a}. Hal\'{i}\v{r}~\textit{et al.}~\cite{halir1998numerically} introduced an improved least squares method to minimize the sum of squared error of the following equation:
\begin{equation}
\label{eq:minimizing}
\begin{aligned}
\min_{\textbf{a}}\sum ^{N}_{i=1} F(x_i,y_i)^2=\min_{\textbf{a}}\sum ^{N}_{i=1} F(\textbf{x}_i)^2\\
\text{where, } \textbf{x}_i=[x_i^2, x_iy_i, y_i^2, x_i, y_i, 1] \text{, and } A_i=(x_i, y_i)
\end{aligned}
\end{equation}
Let us denote the following terms for the fitted ellipse (also see Figure~\ref{fig:ellipse_notaions}):

\begin{labeling}{alligator}
\item [$m$] semi-major axis
\item [$n$] semi-minor axis
\item [$(x_o, y_o)$] center coordinate of the ellipse
\item [$\theta$] rotation angle
\end{labeling}

Be aware that, when the ellipse is near-circular (rotational symmetric shapes), $\theta$ is not stable. A solution for this case is to force $\theta=90^{o}$. However, it did not increase the performance of the VOT datasets empirically.

\subsection{SiamMask\_E}

Since we need to rotate the image with respect to the ellipse center, an affine transformation (Translation and Rotation for our case) will be used here to compute the transformed coordinates. After the estimation of rotation angle $\theta$ and the center point $(x_o, y_o)$, then we need to compute the 2D affine transformation matrix $M$:
\begin{equation}
M
=
\begin{bmatrix}
       cos\Theta & sin\Theta & (1 - cos\Theta) x_o - sin\Theta y_o\\[0.3em]
       -sin\Theta & cos\Theta & sin\Theta x_o - (1 - cos\Theta) y_o
\end{bmatrix}
\end{equation}

Once the affine transformation matrix is computed, then we apply the rotation on the segmentation/mask about the ellipse's center $(x_o, y_o)$:
Let's denote the mask as a set of points $\mathit{Mask}$(magenta color in Figure~\ref{fig:steps}(a)), and the transformed mask as $\mathit{Mask'}$(magenta color in Figure~\ref{fig:steps}(d)).
\begin{equation}
\mathit{Mask'}=M*
\begin{bmatrix}
       x \\[0.3em]
       y \\[0.3em]
       1
\end{bmatrix}
\forall (x, y) \in \mathit{Mask}
\end{equation}

After this step, our aim is to output the intersection (red in Figure~\ref{fig:steps}(f)) between the min-max axis-aligned bounding box (blue in Figure~\ref{fig:steps}(e)) and the ellipse bounding box (green in Figure~\ref{fig:steps}(e)). The advantage of using the ellipse bounding box is to cut out the unexpected portion of the shape (e.g., protruding limbs). Thus, the output bounding box would be able to focus on the trunk of the human body. After the affine transformation, the ellipse bounding box is trivial, and we denote it as $G$:
\begin{equation}
G=[x_o-n, y_o-m, x_o+n, y_o+m]
\end{equation}
The min-max axis-aligned bounding box denote as $B$:
\begin{equation}
\begin{aligned}
B=[&min(\forall x \in \mathit{Mask'}), min(\forall y \in \mathit{Mask'}),\\
   &max(\forall x \in \mathit{Mask'}), max(\forall y \in \mathit{Mask'})]
\end{aligned}
\end{equation}
The intersection bounding box $R$ (red in Figure~\ref{fig:steps}(f)) can be calculated using the following equation:
\begin{equation}
\begin{aligned}
R = [&max(G_1, B_1), max(G_2, B_2),\\
     &min(G_3, B_3), min(G_4, B_4)]
\end{aligned}
\end{equation}
then, convert $R$ to a polygon
\begin{equation}
R = [[R_1,R_2], [R_3, R_2], [R_3, R_4], [R_1, R_4]]
\end{equation}

The last step is to convert the transformed coordinate back to the image coordinate using the inverse of the affine transformation matrix $M$. We denote the output bounding box as $R'$ (red color in Figure~\ref{fig:steps}(g)):
\begin{equation}
R'=M^{-1}*
\begin{bmatrix}
       x \\[0.3em]
       y \\[0.3em]
       1
\end{bmatrix}
;\forall (x, y) \in R
\end{equation}

\subsection{Refinement step (Ref)}
\label{sub:refinement}
\begin{figure}[t]
\begin{center}
\subfigure[Refinement step]{\includegraphics[width = 2.7cm]{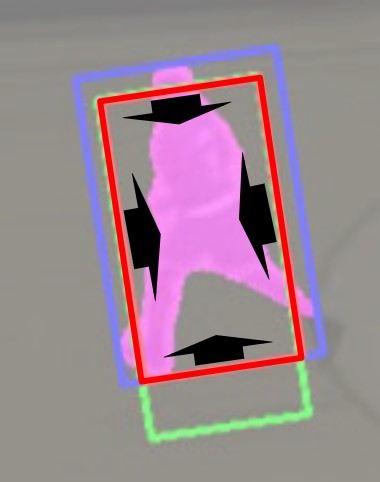}}
\subfigure[Refinement output]{\includegraphics[width = 2.6cm]{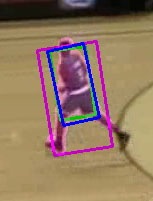}}
\end{center}
\caption{Refinement step: (a) move four edges toward the bounding box center if the constraint in Subsection~\ref{sub:refinement} is not satisfied. (b) the magenta box is the estimated bounding box from SiamMask\_E, and the green box is a sample output after the refinement step. Blue box is the ground truth.}
\label{fig:refinement}
\end{figure}

As you can see from Figure~\ref{fig:siammask_vs_siammaske} at row 3 column 2, our bounding box (green) is not as tight as the ground truth (blue). This problem because the \textit{Mask} generated by SiamMask includes the limbs of the dancer. To manage this problem, we implement a refinement procedure to slim the size of the bounding box by evaluating the amount of \textit{Mask} that an edge is crossing. Let's denote the length of an edge as $\alpha$, and the portion of the edge intersecting the \textit{Mask} is $\beta$. We set a constraint such that, $\beta>\alpha*\textit{factor}$; otherwise, the edge will gradually move toward the bounding box center (see Figure~\ref{fig:refinement}). Here, we choose $\textit{factor}=0.258$ empirically on dataset VOT2018.

%% file: doc/6Experiments.tex
\section{Experiments}
\label{experiments}

\begin{table*}[t]
\begin{center}
\begin{tabular}{l|c|c|c|c|c|c|c|c|c|c}
{} & \multicolumn{3}{c|}{VOT2019} & \multicolumn{3}{c|}{VOT2018} & \multicolumn{3}{c|}{VOT2016} & {} \\
{} & A$\uparrow$ & R$\downarrow$ & EAO$\uparrow$ & A$\uparrow$ & R$\downarrow$ & EAO$\uparrow$ & A$\uparrow$ & R$\downarrow$ & EAO$\uparrow$ & Speed$\uparrow$ \\
\hline
SiamRPN++ & 0.595 & \textbf{0.467} & 0.287 & 0.601 & \textbf{0.234} & 0.415 & 0.642 & \textbf{0.196} & 0.464 & 46 fps  \\
SiamMask & 0.596 & \textbf{0.467} & 0.283 & 0.598 & 0.248 & 0.406 & 0.621 & 0.214 & 0.436 & 87 fps  \\
SiamMask-Opt* & - & - & - & 0.642 & 0.295 & 0.387 & 0.670 & 0.233 & 0.442 & 5 fps  \\
\hline
SiamMask\_E (Ours) & 0.625 & 0.482 & 0.298 & 0.627 & 0.248 & 0.427 & 0.645 & 0.210 & 0.452 & 85 fps \\
SiamMask\_E\_Ref (Ours) & \textbf{0.652} & 0.487 & \textbf{0.309} & \textbf{0.655} & 0.253 & \textbf{0.446} & \textbf{0.677} & 0.224 & \textbf{0.466} & 80 fps
\end{tabular}
\end{center}
\caption{Comparing with the state-of-the-art Siamese trackers on VOT2019, VOT2018, and VOT2016. Our tracker SiamMask\_E with Ref outperforms other trackers in terms of average overlap accuracy (A) and expected average overlap (EAO). $\uparrow$ stands for the higher the best, and $\downarrow$ stands for the lower the best. * the numbers are reported in the original paper.}
\label{tab:vot2016_2018_2019}
\end{table*}

In this section, we evaluate our proposed methods on the datasets that labeled with rotated bounding boxes: VOT2016, VOT2018, and VOT2019.

\subsection{Environment setup}

In order to provide a fair comparison, we test our algorithm using the same pretrained Siamese network model and the same parameters in \cite{wang2018fast}. The reported data is evaluated on a desktop computer with the following hardware:
\begin{itemize}
  \item GPU: GeForce GTX 1080 Ti
  \item CPU: Intel Core i5-8400 CPU @ 2.80GHz $\times$ 6
  \item Memory: 32 GB
\end{itemize}

\subsection{Evaluation metrics}

We only evaluation on the VOT challenge series (VOT2015-2019 short term), where VOT2015 has the same data sequences as VOT2016, and VOT2017 has the same sequences as VOT2018. These three datasets contain 60 sequences with different challenging situations (e.g., motion blur, size change, occlusion, illumination change, etc). To the best of our knowledge, VOT2015-2019 are the only object tracking datasets that labeled with rotated bounding boxes. We also adopt the supervised tracking evaluation methods that are used in VOT2016~\cite{Kristan2016a}: Accuracy (A), Robustness (R), and Expected Average Overlap (EAO). The Accuracy is the average overlap between the estimated and the ground truth bounding boxes when the target is successfully being tracked. The Robustness measures the ratio between the number of times the tracker loses the target (fails) and the number of resumed trackings. The Expected Average Overlap (EAO) is considered as the primary measurement in the VOT challenge. According to the official toolkit, the tracker will be reinitialized when the estimated bounding box has no intersection with the ground truth. After five frames, the tracker will restart with the ground truth bounding box.

\subsection{Overall results}

Table~\ref{tab:vot2016_2018_2019} presents the result comparison between the state-of-the-art Siamese based tracking algorithms on VOT2016, VOT2018, and VOT2019 datasets. Our tracker SiamMask\_E with Ref has the 0.655 Accuracy and 0.446 EAO on VOT2018 dataset which it a new state-of-the-art comparing the other Siamese trackers and the VOT2018 short term challenge winners~\cite{Kristan2018a}. Although SiamMask-Opt has the similar performance as ours, due to the computation complexity, SiamMask-Opt can only run at 5 frames per second. However, our tracker is able to process in real-time with the speed of more than 80 frames per second. Similarly, our tracker also forms a new state-of-the-art result on VOT2019.



\subsection{Ablation studies}

\begin{table*}[t]
\begin{center}
\begin{tabular}{l|c|c|c|c|c|c}
{} & \multicolumn{3}{c|}{VOT2019} & \multicolumn{3}{c}{VOT2018}\\
{} & A$\uparrow$ & R$\downarrow$ & EAO$\uparrow$ & A$\uparrow$ & R$\downarrow$ & EAO$\uparrow$\\
 \hline
 SiamMask\_E  (Ours) & 0.625 & 0.482 & 0.298 & 0.627 & 0.248 & 0.427 \\
 SiamMask            & 0.596 & 0.467 & 0.283 & 0.598 & 0.248 & 0.406 \\
  \hline
 SiamMask\_E + minABoxAngle & 0.618 & 0.472 & 0.292 & 0.621 & 0.253 & 0.418 \\
 SiamMask + ellipseAngle         & 0.594 & 0.477 & 0.284 & 0.595 & 0.243 & 0.409 \\
 \hline\hline
 SiamMask\_E + Ref (Ours) & \textbf{0.652} & 0.487 & \textbf{0.309} & \textbf{0.655} & 0.253 & \textbf{0.446} \\
 SiamMask + Ref           & 0.644 & 0.497 & 0.297 & 0.645 & 0.272 & 0.416 \\
 \hline
 SiamMask\_E + minABoxAngle + Ref & 0.647 & 0.487 & 0.302 & 0.649 & 0.262 & 0.428 \\
 SiamMask + ellipseAngle + Ref    & 0.639 & 0.502 & 0.299 & 0.643 & 0.267 & 0.422 \\
 \hline
\end{tabular}
\end{center}
\caption{Ablation studies: SiamMask\_E is our baseline tracker with the ellipse angle and ellipse box, and SiamMask is the original tracker with the minimum area bounding box. Ref stands for the refinement step in Subsection~\ref{sub:refinement}. minABoxAngle stands for the orientation of the minimum area bounding box. ellipseAngle stands for the orientation of the best fitting ellipse. The result shows that the effectiveness of ellipse orientation and refinement step significantly improve the performance of SiamMask.}
\label{tab:ablation_test}
\end{table*}

The ablation test results are shown in Table~\ref{tab:ablation_test}. In the table, SiamMask\_E is our baseline model without the refinement step. We exchange the bounding box orientation between SiamMask\_E and SiamMask, where SiamMask\_E with Minimum Area Bounding Box angle (SiamMask\_E + minABoxAngle) performs weaker than our baseline SiamMask\_E. Similarly, SiamMask with ellipse angle (SiamMask + ellipseAngle) is preferable over the original SiamMask. 
By adding the refinement step (Ref) to both SiamMask and SiamMask\_E, the average overlap Accuracy increase dramatically. 
Furthermore, we modify the bounding box rotation of SiamMask\_E + Ref with the original Minimum Area Bounding Box angle ( SiamMask\_E + Ref + minABoxAngle) which results in slightly decreasing on the primary measurement EAO. It proves that using the ellipse's angle could improve the tracking performance on the VOT datasets. 
On the other hand, we also test SiamMask + Ref with changing the angle of the Minimum Area Bounding Box to the ellipse's angle (SiamMask + ellipseAngle + Ref). The result shows that SiamMask + ellipseAngle + Ref also has some degree of improvement on both VOT2018 and VOT2019 on the primary measurement EAO. Overall, SiamMask\_E, which improves the bounding box orientation and scale using ellipse fitting on top of SiamMask, has a similar performance as the original SimaMask with the refinement step (SiamMask + Ref). And, SiamMask\_E with the refinement step (SiamMask\_E + Ref) outperforms any other combinations on the ablation study table.

\subsection{Qualitative results}

\begin{figure*}[t]
\begin{center}
\newcommand{\inputimagewidth}{2.6cm}
\setlength{\tabcolsep}{0.15em} 
\begin{tabular}{ccccccccc}
\rotatebox{90}{basketball}&
\includegraphics[width = \inputimagewidth]{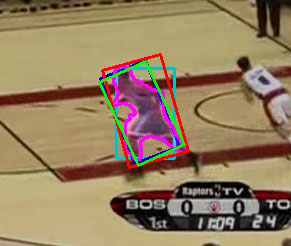}&
\includegraphics[width = \inputimagewidth]{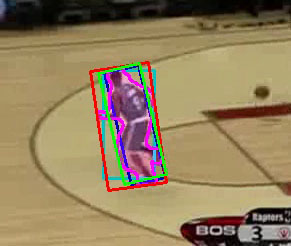}&
\includegraphics[width = \inputimagewidth]{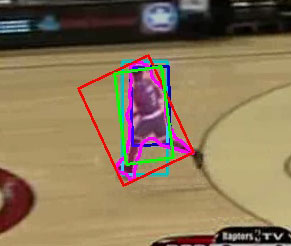}&
\includegraphics[width = \inputimagewidth]{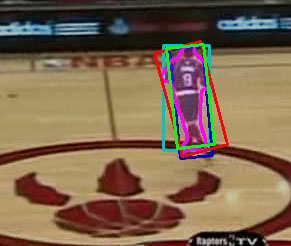}&
\includegraphics[width = \inputimagewidth]{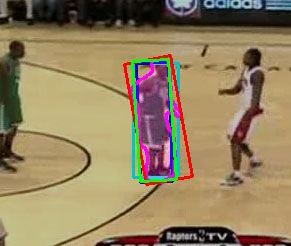}&
\includegraphics[width = \inputimagewidth]{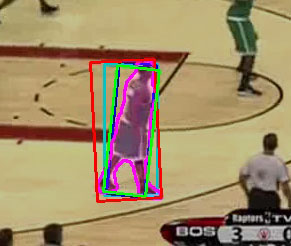}\\
\rotatebox{90}{basketball}&
\includegraphics[width = \inputimagewidth]{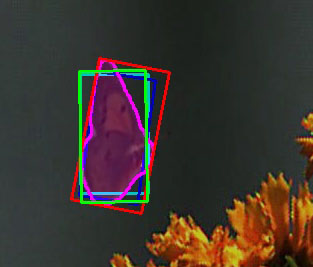}&
\includegraphics[width = \inputimagewidth]{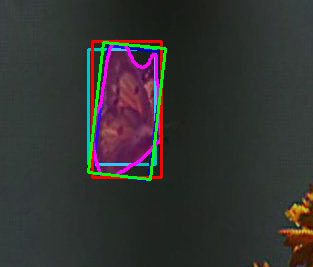}&
\includegraphics[width = \inputimagewidth]{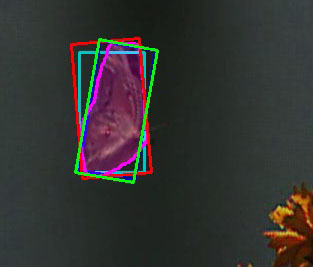}&
\includegraphics[width = \inputimagewidth]{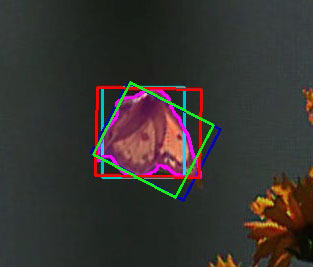}&
\includegraphics[width = \inputimagewidth]{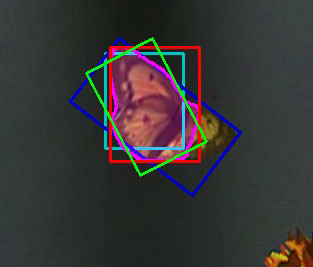}&
\includegraphics[width = \inputimagewidth]{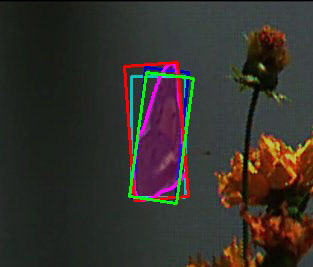}\\
\rotatebox{90}{fish1}&
\includegraphics[width = \inputimagewidth]{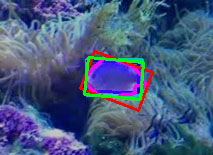}&
\includegraphics[width = \inputimagewidth]{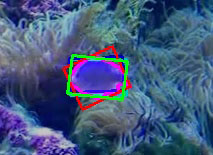}&
\includegraphics[width = \inputimagewidth]{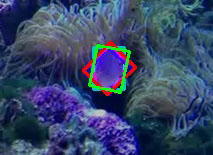}&
\includegraphics[width = \inputimagewidth]{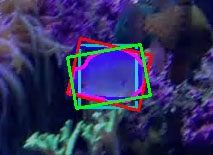}&
\includegraphics[width = \inputimagewidth]{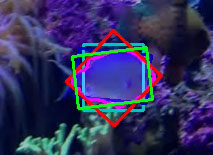}&
\includegraphics[width = \inputimagewidth]{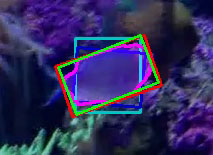}\\
\rotatebox{90}{graduate}&
\includegraphics[width = \inputimagewidth]{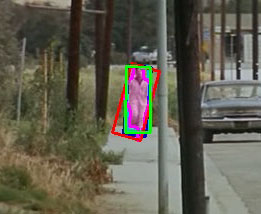}&
\includegraphics[width = \inputimagewidth]{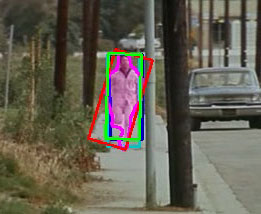}&
\includegraphics[width = \inputimagewidth]{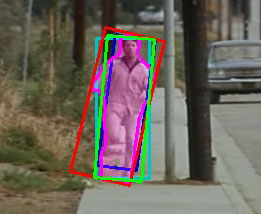}&
\includegraphics[width = \inputimagewidth]{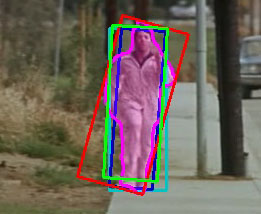}&
\includegraphics[width = \inputimagewidth]{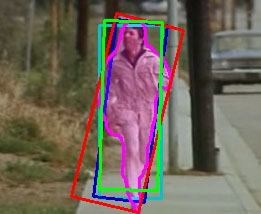}&
\includegraphics[width = \inputimagewidth]{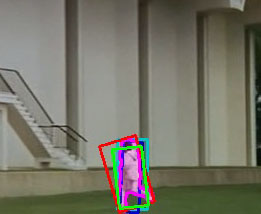}\\
\rotatebox{90}{iceskater1}&
\includegraphics[width = \inputimagewidth]{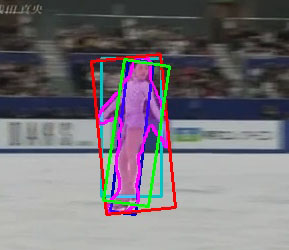}&
\includegraphics[width = \inputimagewidth]{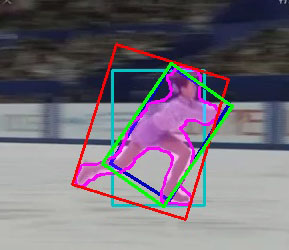}&
\includegraphics[width = \inputimagewidth]{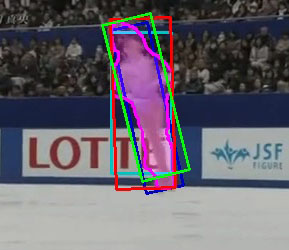}&
\includegraphics[width = \inputimagewidth]{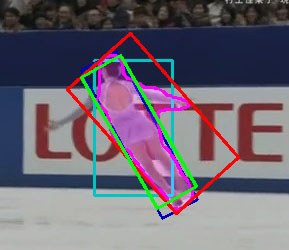}&
\includegraphics[width = \inputimagewidth]{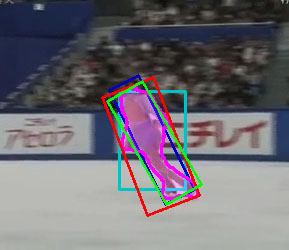}&
\includegraphics[width = \inputimagewidth]{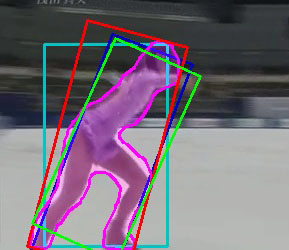}\\
\rotatebox{90}{monkey}&
\includegraphics[width = \inputimagewidth]{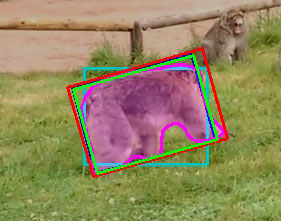}&
\includegraphics[width = \inputimagewidth]{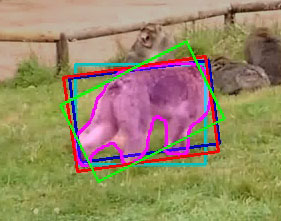}&
\includegraphics[width = \inputimagewidth]{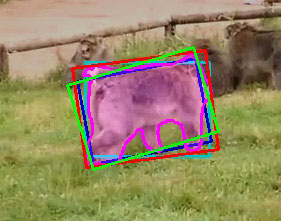}&
\includegraphics[width = \inputimagewidth]{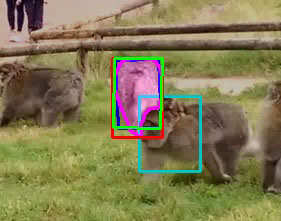}&
\includegraphics[width = \inputimagewidth]{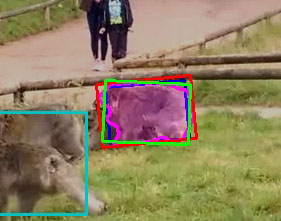}&
\includegraphics[width = \inputimagewidth]{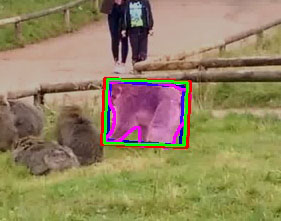}\\
\rotatebox{90}{polo}&
\includegraphics[width = \inputimagewidth]{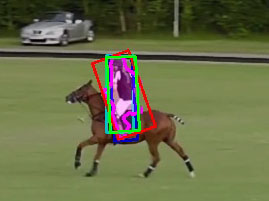} &
\includegraphics[width = \inputimagewidth]{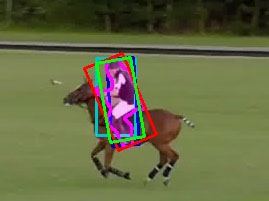} &
\includegraphics[width = \inputimagewidth]{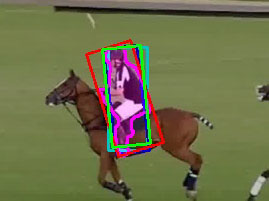} &
\includegraphics[width = \inputimagewidth]{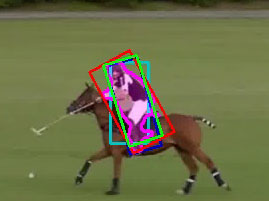} &
\includegraphics[width = \inputimagewidth]{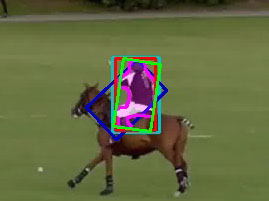} &
\includegraphics[width = \inputimagewidth]{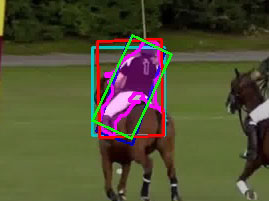} \\
\rotatebox{90}{surfing}&
\includegraphics[width = \inputimagewidth]{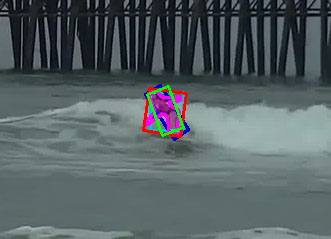} &
\includegraphics[width = \inputimagewidth]{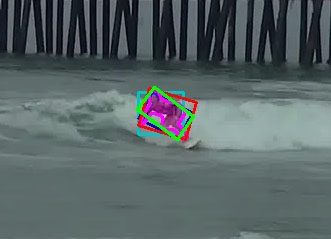} &
\includegraphics[width = \inputimagewidth]{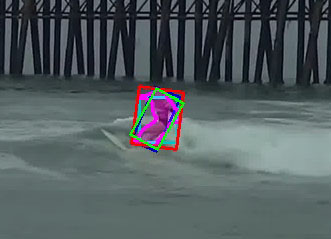} &
\includegraphics[width = \inputimagewidth]{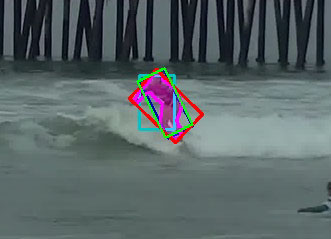} &
\includegraphics[width = \inputimagewidth]{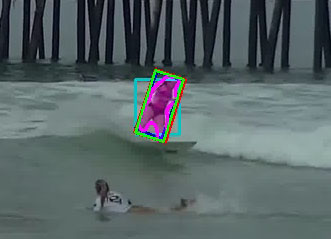} &
\includegraphics[width = \inputimagewidth]{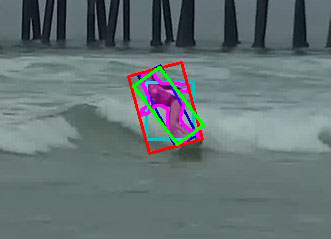} \\
\end{tabular}
\end{center}
\caption{Qualitative results: We show some sample outputs on eight sequences selected from VOT2019~\cite{Kristan2019a}, where the red box is SiamMask~\cite{wang2018fast}, the cyan box is SiamRPN++~\cite{li2018siamrpn++}, the green box is SiamMask\_E(ours), and the blue box is the ground truth.}
\label{fig:display_cases}
\end{figure*}

To analysis the improvement, we show several results computed on VOT2019~\cite{Kristan2019a} dataset. We compare the state-of-the-art algorithms SiamMask~\cite{wang2018fast} and SiamRPN++~\cite{li2018siamrpn++} along with our approach SiamMask\_E in Figure~\ref{fig:display_cases}.

%% file: doc/7Conclusion.tex
\section{Conclusion}
\label{conclusion}

In this paper, we updated the \textbf{SiamMask} tracker to achieve the next level of state-of-the-art performance. Our new tracker \textbf{SiamMask\_E} retains real-time processing speed at 80 fps. We show that the bounding box using ellipse fitting outperforms the minimum area rectangle bounding box in terms of better rotation angle and tighter bounding box scale. Our results show the strength of SiamMask network tracking model such that it can outperform the other state-of-the-art trackers.

\textbf{Future work}: Our approach focused on an efficient bounding box refinement algorithm. On a different aspect, if a proper motion model is employed, we believe the result could move to the next level. To attain this, a real-time algorithm is needed to differentiate the camera the target motion in order to estimate the real target motion. As well as, we need to beware the other dynamic distractors in the scene. 

%% file: doc/8Acknowledgement.tex
\ificcvfinal\section*{Acknowledgement}
We acknowledge the financial support of the Natural Sciences and Engineering Research Council of Canada (NSERC), the NSERC Canadian Robotics Network (NCRN), and the Canada Research Chairs Program through grants to John K. Tsotsos. The fist author also would like to thank Dekun Wu provided hardware support for testing.
\fi

%% file: final.bbl
\begin{thebibliography}{10}\itemsep=-1pt

\bibitem{agarwal2018study}
N.~Agarwal, C.-W. Chiang, and A.~Sharma.
\newblock A study on computer vision techniques for self-driving cars.
\newblock In {\em International Conference on Frontier Computing}, pages
  629--634. Springer, 2018.

\bibitem{bertinetto2016fully}
L.~Bertinetto, J.~Valmadre, J.~F. Henriques, A.~Vedaldi, and P.~H. Torr.
\newblock Fully-convolutional siamese networks for object tracking.
\newblock In {\em European conference on computer vision}, pages 850--865.
  Springer, 2016.

\bibitem{brox2010large}
T.~Brox and J.~Malik.
\newblock Large displacement optical flow: descriptor matching in variational
  motion estimation.
\newblock {\em IEEE transactions on pattern analysis and machine intelligence},
  33(3):500--513, 2010.

\bibitem{buyval2018realtime}
A.~Buyval, A.~Gabdullin, R.~Mustafin, and I.~Shimchik.
\newblock Realtime vehicle and pedestrian tracking for didi udacity
  self-driving car challenge.
\newblock In {\em 2018 IEEE International Conference on Robotics and Automation
  (ICRA)}, pages 2064--2069. IEEE, 2018.

\bibitem{chen2017integrating}
B.~X. Chen, R.~Sahdev, and J.~K. Tsotsos.
\newblock Integrating stereo vision with a cnn tracker for a person-following
  robot.
\newblock In {\em International Conference on Computer Vision Systems}, pages
  300--313. Springer, 2017.

\bibitem{chen2017personfollowing}
B.~X. Chen, R.~Sahdev, and J.~K. Tsotsos.
\newblock Person following robot using selected online ada-boosting with stereo
  camera.
\newblock In {\em Computer and Robot Vision (CRV), 2017 14th Conference on},
  pages 48--55. IEEE, 2017.

\bibitem{cho2014multi}
H.~Cho, Y.-W. Seo, B.~V. Kumar, and R.~R. Rajkumar.
\newblock A multi-sensor fusion system for moving object detection and tracking
  in urban driving environments.
\newblock In {\em 2014 IEEE International Conference on Robotics and Automation
  (ICRA)}, pages 1836--1843. IEEE, 2014.

\bibitem{fitzgibbon1996buyer}
A.~W. Fitzgibbon, R.~B. Fisher, et~al.
\newblock {\em A buyer's guide to conic fitting}.
\newblock University of Edinburgh, Department of Artificial Intelligence, 1996.

\bibitem{gajjar2017human}
V.~Gajjar, A.~Gurnani, and Y.~Khandhediya.
\newblock Human detection and tracking for video surveillance: A cognitive
  science approach.
\newblock In {\em Proceedings of the IEEE International Conference on Computer
  Vision}, pages 2805--2809, 2017.

\bibitem{gander1980least}
W.~Gander.
\newblock Least squares with a quadratic constraint.
\newblock {\em Numerische Mathematik}, 36(3):291--307, 1980.

\bibitem{guo2017structure}
Q.~Guo, W.~Feng, C.~Zhou, C.-M. Pun, and B.~Wu.
\newblock Structure-regularized compressive tracking with online data-driven
  sampling.
\newblock {\em IEEE Transactions on Image Processing}, 26(12):5692--5705, 2017.

\bibitem{halir1998numerically}
R.~Hal{\i}r and J.~Flusser.
\newblock Numerically stable direct least squares fitting of ellipses.
\newblock In {\em Proc. 6th International Conference in Central Europe on
  Computer Graphics and Visualization. WSCG}, volume~98, pages 125--132.
  Citeseer, 1998.

\bibitem{he2018towards}
A.~He, C.~Luo, X.~Tian, and W.~Zeng.
\newblock Towards a better match in siamese network based visual object
  tracker.
\newblock In {\em Proceedings of the European Conference on Computer Vision
  (ECCV)}, pages 0--0, 2018.

\bibitem{he2018twofold}
A.~He, C.~Luo, X.~Tian, and W.~Zeng.
\newblock A twofold siamese network for real-time object tracking.
\newblock In {\em Proceedings of the IEEE Conference on Computer Vision and
  Pattern Recognition}, pages 4834--4843, 2018.

\bibitem{henriques2014high}
J.~F. Henriques, R.~Caseiro, P.~Martins, and J.~Batista.
\newblock High-speed tracking with kernelized correlation filters.
\newblock {\em IEEE transactions on pattern analysis and machine intelligence},
  37(3):583--596, 2014.

\bibitem{hough1962method}
P.~V. Hough.
\newblock Method and means for recognizing complex patterns, Dec.~18 1962.
\newblock US Patent 3,069,654.

\bibitem{hua2015online}
Y.~Hua, K.~Alahari, and C.~Schmid.
\newblock Online object tracking with proposal selection.
\newblock In {\em Proceedings of the IEEE international conference on computer
  vision}, pages 3092--3100, 2015.

\bibitem{koide2018convolutional}
K.~Koide and J.~Miura.
\newblock Convolutional channel features-based person identification for person
  following robots.
\newblock In {\em International Conference on Intelligent Autonomous Systems},
  pages 186--198. Springer, 2018.

\bibitem{Kristan2018a}
M.~Kristan, A.~Leonardis, J.~Matas, M.~Felsberg, R.~Pflugfelder, L.~\v{C}ehovin
  Zajc, T.~Vojir, G.~H\"{a}ger, A.~Luke\v{z}i\v{c}, A.~Eldesokey, G.~Fernandez,
  and et~al.
\newblock The sixth visual object tracking vot2018 challenge results, 2018.

\bibitem{Kristan2019a}
M.~Kristan, A.~Leonardis, J.~Matas, M.~Felsberg, R.~Pflugfelder, L.~\v{C}ehovin
  Zajc, T.~Vojir, G.~H\"{a}ger, A.~Luke\v{z}i\v{c}, A.~Eldesokey, G.~Fernandez,
  and et~al.
\newblock The seventh visual object tracking vot2019 challenge results, 2019.

\bibitem{Kristan2016a}
M.~Kristan, A.~Leonardis, J.~Matas, M.~Felsberg, R.~Pflugfelder, L.~\v{C}ehovin
  Zajc, T.~Vojir, G.~H\"{a}ger, A.~Luke\v{z}i\v{c}, and G.~Fernandez.
\newblock The visual object tracking vot2016 challenge results.
\newblock Springer, Oct 2016.

\bibitem{krizhevsky2012imagenet}
A.~Krizhevsky, I.~Sutskever, and G.~E. Hinton.
\newblock Imagenet classification with deep convolutional neural networks.
\newblock In {\em Advances in neural information processing systems}, pages
  1097--1105, 2012.

\bibitem{lee2017online}
Y.-G. Lee, Z.~Tang, and J.-N. Hwang.
\newblock Online-learning-based human tracking across non-overlapping cameras.
\newblock {\em IEEE Transactions on Circuits and Systems for Video Technology},
  28(10):2870--2883, 2017.

\bibitem{li2018siamrpn++}
B.~Li, W.~Wu, Q.~Wang, F.~Zhang, J.~Xing, and J.~Yan.
\newblock Siamrpn++: Evolution of siamese visual tracking with very deep
  networks.
\newblock {\em arXiv preprint arXiv:1812.11703}, 2018.

\bibitem{li2018high}
B.~Li, J.~Yan, W.~Wu, Z.~Zhu, and X.~Hu.
\newblock High performance visual tracking with siamese region proposal
  network.
\newblock In {\em Proceedings of the IEEE Conference on Computer Vision and
  Pattern Recognition}, pages 8971--8980, 2018.

\bibitem{lin2014microsoft}
T.-Y. Lin, M.~Maire, S.~Belongie, J.~Hays, P.~Perona, D.~Ramanan,
  P.~Doll{\'a}r, and C.~L. Zitnick.
\newblock Microsoft coco: Common objects in context.
\newblock In {\em European conference on computer vision}, pages 740--755.
  Springer, 2014.

\bibitem{nam2016learning}
H.~Nam and B.~Han.
\newblock Learning multi-domain convolutional neural networks for visual
  tracking.
\newblock In {\em Proceedings of the IEEE Conference on Computer Vision and
  Pattern Recognition}, pages 4293--4302, 2016.

\bibitem{nebehay2014consensus}
G.~Nebehay and R.~Pflugfelder.
\newblock Consensus-based matching and tracking of keypoints for object
  tracking.
\newblock In {\em IEEE Winter Conference on Applications of Computer Vision},
  pages 862--869. IEEE, 2014.

\bibitem{nebehay2015clustering}
G.~Nebehay and R.~Pflugfelder.
\newblock Clustering of static-adaptive correspondences for deformable object
  tracking.
\newblock In {\em Proceedings of the IEEE Conference on Computer Vision and
  Pattern Recognition}, pages 2784--2791, 2015.

\bibitem{petrovskaya2009model}
A.~Petrovskaya and S.~Thrun.
\newblock Model based vehicle detection and tracking for autonomous urban
  driving.
\newblock {\em Autonomous Robots}, 26(2-3):123--139, 2009.

\bibitem{ren2016real}
Q.~Ren, Q.~Zhao, H.~Qi, and L.~Li.
\newblock Real-time target tracking system for person-following robot.
\newblock In {\em 2016 35th Chinese Control Conference (CCC)}, pages
  6160--6165. IEEE, 2016.

\bibitem{ren2015faster}
S.~Ren, K.~He, R.~Girshick, and J.~Sun.
\newblock Faster r-cnn: Towards real-time object detection with region proposal
  networks.
\newblock In {\em Advances in neural information processing systems}, pages
  91--99, 2015.

\bibitem{rout2018rotation}
L.~Rout, D.~Mishra, R.~K. S.~S. Gorthi, et~al.
\newblock Rotation adaptive visual object tracking with motion consistency.
\newblock In {\em 2018 IEEE Winter Conference on Applications of Computer
  Vision (WACV)}, pages 1047--1055. IEEE, 2018.

\bibitem{russakovsky2015imagenet}
O.~Russakovsky, J.~Deng, H.~Su, J.~Krause, S.~Satheesh, S.~Ma, Z.~Huang,
  A.~Karpathy, A.~Khosla, M.~Bernstein, et~al.
\newblock Imagenet large scale visual recognition challenge.
\newblock {\em International journal of computer vision}, 115(3):211--252,
  2015.

\bibitem{wang2018fast}
Q.~Wang, L.~Zhang, L.~Bertinetto, W.~Hu, and P.~H. Torr.
\newblock Fast online object tracking and segmentation: A unifying approach.
\newblock {\em arXiv preprint arXiv:1812.05050}, 2018.

\bibitem{WuLimYang13}
Y.~Wu, J.~Lim, and M.-H. Yang.
\newblock Online object tracking: A benchmark.
\newblock In {\em IEEE Conference on Computer Vision and Pattern Recognition
  (CVPR)}, 2013.

\bibitem{WuLimYang15}
Y.~Wu, J.~Lim, and M.-H. Yang.
\newblock Object tracking benchmark.
\newblock In {\em IEEE Transactions on Pattern Analysis and Machine
  Intelligence (TPAMI)}, volume~37, pages 1834--1848, 2015.

\bibitem{xu2018youtube}
N.~Xu, L.~Yang, Y.~Fan, J.~Yang, D.~Yue, Y.~Liang, B.~Price, S.~Cohen, and
  T.~Huang.
\newblock Youtube-vos: Sequence-to-sequence video object segmentation.
\newblock In {\em Proceedings of the European Conference on Computer Vision
  (ECCV)}, pages 585--601, 2018.

\bibitem{xu2018real}
R.~Xu, S.~Y. Nikouei, Y.~Chen, A.~Polunchenko, S.~Song, C.~Deng, and T.~R.
  Faughnan.
\newblock Real-time human objects tracking for smart surveillance at the edge.
\newblock In {\em 2018 IEEE International Conference on Communications (ICC)},
  pages 1--6. IEEE, 2018.

\bibitem{zhang2015joint}
M.~Zhang, J.~Xing, J.~Gao, X.~Shi, Q.~Wang, and W.~Hu.
\newblock Joint scale-spatial correlation tracking with adaptive rotation
  estimation.
\newblock In {\em Proceedings of the IEEE International Conference on Computer
  Vision Workshops}, pages 32--40, 2015.

\bibitem{zhou2016moving}
Y.~Zhou, S.~Zlatanova, Z.~Wang, Y.~Zhang, and L.~Liu.
\newblock Moving human path tracking based on video surveillance in 3d indoor
  scenarios.
\newblock {\em ISPRS Annals of the Photogrammetry, Remote Sensing and Spatial
  Information Sciences}, 3:97, 2016.

\end{thebibliography}
